\documentclass{bmvc2k}

\title{SPG-Net: Segmentation Prediction and Guidance Network for Image Inpainting}

\addauthor{Yuhang Song}{yuhangso@usc.edu}{1}
\addauthor{Chao Yang}{chaoy@usc.edu}{1}
\addauthor{Yeji Shen}{yejishen@usc.edu}{1}
\addauthor{Peng Wang}{wangpeng54@baidu.com}{2}
\addauthor{Qin Huang}{qinhuang@usc.edu}{1}
\addauthor{C.-C. Jay Kuo}{cckuo@sipi.usc.edu}{1}

\addinstitution{
 University of Southern California\\
 3740 McClintock Ave\\
 Los Angeles, USA
}
\addinstitution{
 Baidu Research\\
 1195 Bordeaux Dr.,\\
 Sunnyvale, USA
}

\runninghead{Song \emph{et al}\bmvaOneDot}{Segmentation-Guided Image Inpainting}

\def\eg{\emph{e.g}\bmvaOneDot}

\def\etal{\emph{et al}\bmvaOneDot}

\begin{document}

\maketitle

\begin{abstract}

In this paper, we focus on image inpainting task, aiming at recovering the missing area of an incomplete image given the context information. Recent development in deep generative models enables an efficient end-to-end framework for image synthesis and inpainting tasks, but existing methods based on generative models don't exploit the segmentation information to constrain the object shapes, which usually lead to blurry results on the boundary. To tackle this problem, we propose to introduce the semantic segmentation information, which disentangles the inter-class difference and intra-class variation for image inpainting. This leads to much clearer recovered boundary between semantically different regions and better texture within semantically consistent segments. Our model factorizes the image inpainting process into segmentation prediction (SP-Net) and segmentation guidance (SG-Net) as two steps, which predict the segmentation labels in the missing area first, and then generate segmentation guided inpainting results. Experiments on multiple public datasets show that our approach outperforms existing methods in optimizing the image inpainting quality, and the interactive segmentation guidance provides possibilities for multi-modal predictions of image inpainting.

\end{abstract}

\section{Introduction}
\label{sec:introduction}

\begin{figure}[!h]
\centering
\small
\setlength{\tabcolsep}{1pt}
\begin{tabular}{ccccc}
  \includegraphics[width=.20\textwidth]{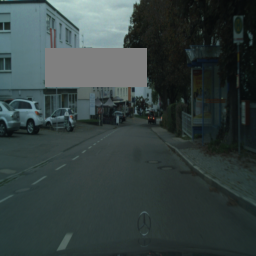}&
  \includegraphics[width=.20\textwidth]{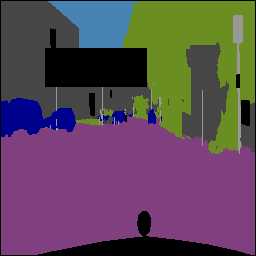}&
  \includegraphics[width=.20\textwidth]{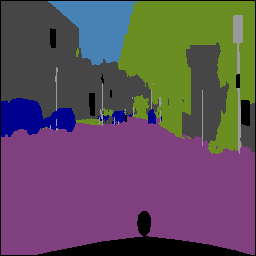}&
  \includegraphics[width=.20\textwidth]{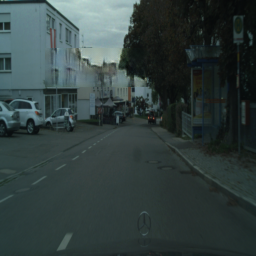}&
  \includegraphics[width=.20\textwidth]{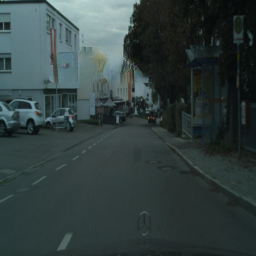}\\
(a) & (b) & (c) & (d) & (e)\\
\end{tabular}
\caption{Comparison of our intermediate and final result with GL inpainting~\cite{IizukaSIGGRAPH2017}. (a) Input image with missing hole. (b) Deeplabv3+~\cite{deeplabv3plus2018} output. (c) SP-Net result. (d) SG-Net result. (e) Inpainting result given by GL inpainting~\cite{IizukaSIGGRAPH2017}. The size of images are 256x256.}
\vspace{-10pt}
\label{fig:teaser}
\end{figure}

Image inpainting is the task to reconstruct the missing region in an image with plausible contents based on its surrounding context, which is a common topic of low-level computer vision~\cite{hays2007scene,komodakis2006image}. Making use of this technique, people could restore damaged images or remove unwanted objects from images or videos. In this task, our goal is to not only fill in the plausible contexts with realistic details but also make the inpainted area coherent with the contexts as well as the boundaries. 

Traditional image inpainting methods mostly use image-level features to address the problem of filling in the hole. A typical method is Patch-Match~\cite{barnes2009patchmatch}, in which Barnes \etal which proposes to search for the best matching patches to reconstruct the missing area. Another example is ~\cite{wilczkowiak2005hole}, which further optimizes the search areas and find the most fitting patches. These methods could provide realistic texture by its nature, however, they only make use of the low-level features of the given context and lack the ability to predict the high-level features in the missing hole. On the other hand, instead of capturing the global structure of the images, they propagate the texture from outside into the hole. This often leads to semantically inconsistent inpainting results which are unwanted by humans.

Recent developments of deep generative models have enabled the generation of realistic images either from noise vectors or conditioned on some prior knowledge, such as images, labels, or word embeddings. In this way, we could regard the image inpainting task as an image generation task conditioned on the given context of images~\cite{pathak2016context, yang2017high,iizuzuka2017globally,li2017generative,yeh2017semantic}. One of the earliest works that apply the deep generative model to image inpainting task is Context-encoder~\cite{pathak2016context}, where Pathak \etal trains an encoder-decoder architecture to predict the complete image directly from the input image with a hole. Adding the adversarial loss has enabled large improvement on the image inpainting quality, but the results still lack high-frequency details and contain notable artifacts.

To handle higher resolution inpainting problems, Iizuka \etal~\cite{iizuzuka2017globally} proposes to add dilation convolution layers to increase the receptive field and use a joint global and local discriminator to improve the consistency of the image completion result. However, their results often contain noise patterns and artifacts which need to be reduced by a post-processing step (\eg Poisson image editing~\cite{perez2003poisson}). Meanwhile, the training of their method is very time-consuming, which takes around 2 months in total. Another line of work in high-resolution inpainting is trying to apply style transfer methods to refine the inpainting texture. More specifically, Yang \etal~\cite{yang2017high} proposes to optimize the inpainting result by finding the best matching neural patches between the inpainting area and the given context, and then a multi-scale structure is applied to refine the texture in an iterative way to achieve the high-resolution performance. It could predict photo-realistic results but the inference takes much more time than other methods.

Another limitation of many recent approaches is that they usually predict the complete images directly and don't exploit the segmentation information from the images. We find that this limitation usually leads to blurry boundaries between different objects in the inpainting area. To address this problem, we propose to use segmentation mask as an intermediate bridge for the incomplete image and the complete image prediction. We decouple the inpainting process into two steps: segmentation prediction (SP-Net) and segmentation guidance (SG-Net). We first use a state-of-the-art image segmentation method~\cite{deeplabv3plus2018} to generate the segmentation labels for the input image. Then we predict the segmentation label in the missing area directly, which gives us a prior knowledge of predicted object localization and shape details in the hole. Finally we combine this complete segmentation mask with the input image together and pass them into the segmentation guidance network to make the complete prediction. This formulates the segmentation guided semantic segmentation process (see Fig.~\ref{fig:teaser}), and the whole system is able to combine the strength of deep generative models as well as the segmentation information, which can guide the architecture to make a more realistic prediction, especially for boundaries between different objects. On the other hand, as compared with other methods which could only make a single prediction given the input image, our method provides the possibility of interactive and multi-modal predictions. More specifically, users could edit the segmentation mask in the missing hole interactively, and the predictions could be different according to assigned segmentation labels in the hole. 

We evaluate the performance of the proposed framework on a variety of datasets based on both qualitative and quantitative evaluations. We also provide a thorough analysis and ablation study about different steps in our architecture. The experimental results demonstrate that the segmentation map offers useful information in generating texture details, which leads to better image inpainting quality.

The rest of this paper is organized as follows. The related work is reviewed in Section \ref{sec:relatedwork}. The full architecture and the approach are proposed in Section \ref{sec:approach}. The experimental results and evaluations are presented in Section \ref{sec:experiments}. Finally, the conclusion is given in Section \ref{sec:conclusion}.


\section{Related Work}
\label{sec:relatedwork}

\begin{figure*}[t]
	\centering
	\includegraphics[width=.98\linewidth]{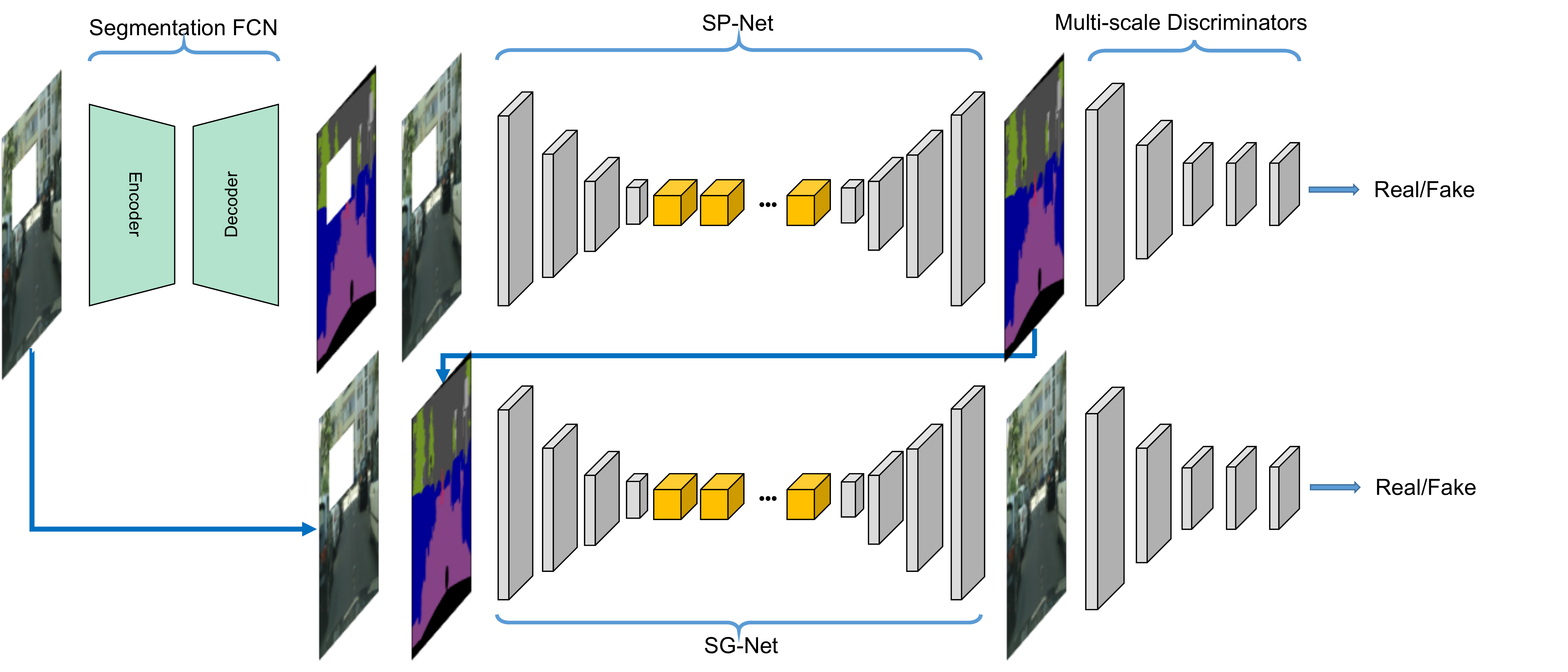}
	\caption{Overview of our network architecture. We firstly use Deeplabv3+ for the image segmentation initialization, and then use SP-Net to predict the segmentation mask. Finally, the segmentation mask is combined with the original input image to go through the SG-Net which generates the complete image prediction.} 
	\vspace{-10pt}
\end{figure*}
\label{fig:framework}

Traditional image inpainting approaches make use of the image-level features to propagate the texture from the surrounding context to the missing hole~\cite{bertalmio2000image, ballester2001filling}. These methods can only tackle small holes and would lead to obvious artifacts and noise patterns for large holes. Later works using patch-based methods~\cite{kwatra2005texture, barnes2009patchmatch} could optimize the inpainting performance by searching the best matching patches. However, while these methods could provide plausible texture generation in the hole, they are not aware of the semantic structure of the image and cannot make reasonable inference for object completion and structure prediction.

Deep learning base methods benefit a lot from recent developments of deep generative models, especially the generative adversarial networks (GANs)~\cite{goodfellow2014generative}, which implicitly model the density, and show promising performance to generate visually realistic images. However, it is quite difficult to make the training of GANs stable and to generate high resolution images. To this end, extensive works are proposed to stabilize the training process, such as DC-GAN~\cite{radford2015unsupervised}, Wasserstein GAN (WGAN)~\cite{salimans2016improved,arjovsky2017wasserstein}, WGAN-GP~\cite{gulrajani2017improved}, LSGAN~\cite{mao2016least} and Progressive GANs~\cite{karras2017progressive}. These methods provide practical techniques for training GANs and enable photo-realistic high-resolution image generation and synthesis. 

With the rapid development of adversarial training, Context-encoder~\cite{pathak2016context} introduces adversarial loss to the image inpainting task, which encodes the 128x128 incomplete image to low dimension feature and then decodes it back to the image space using a joint loss of reconstruction loss and adversarial loss. In~\cite{yeh2017semantic}, Yeh \etal proposes to search for the closest matching of the corrupted image in the feature space to reconstruct the missing area. Iizuka \etal~\cite{iizuzuka2017globally} improves the work by defining both global and local discriminators with a post-processing step, which achieves good performance but is hard to train. In ~\cite{ulyanov2017deep}, Ulyanov \etal points out that the structure of a generator network itself is sufficient to capture the low-level image statistics and could restore images based on the image prior without additional training. In ~\cite{yang2018image}, Yang \etal designs a block-wise procedural traingin scheme and an adversarial loss annealing strategy to stabilize the training. In~\cite{yu2018generative}, Yu \etal adds a contextual attention layer to match the feature patches from given surronding backgrounds. In~\cite{liu2018image}, Liu \etal proposes to use a partial convolution layer with an automatic mask updating mechanism to reduce artifacts for irregular tasks. Another class of methods focuses on guiding the texture synthesis procedure with a prior content initialization. In~\cite{yang2017high}, Yang \etal proposes to predict an initial low-resolution result first and then optimize the synthesized texture by propagating the fine texture from surrounding context to the inside of the hole. In~\cite{song2017image} Song \etal adds a texture refinement network to iteratively optimize the texture by matching the closest patches in the latent space. 

In comparison with previous related work, we propose a multi-network system that addresses the inpainting of segmentation labels and images simultaneously. The topic of semantic segmentation has been extensively researched since the emerging of Fully Convolutional Networks (FCNs) \cite{sermanet2013overfeat, long2015fully}. The encoder-decoder models~\cite{badrinarayanan2017segnet,ronneberger2015u,lin2017refinenet} could reduce the spatial dimension first to capture the global information and then recover the object details. In ~\cite{zheng2015conditional, vemulapalli2016gaussian}, they propose to incorporate an extra Conditional Random Fields module to encode long-range context. In ~\cite{liang2015semantic, deeplabv3plus2018}, they propose to use atrous convolutional layers to capture multi-scale context. In this paper, we propose to regard the semantic segmentation labels as the latent variables and use them as the hint to guide the image inpainting task. It is shown that this strategy could produce sharper and clearer texture especially between boundaries of different objects. Meanwhile, an interactive segmentation guidance could be made by users to generate multi-modal inpainting results.

\section{Approach}\label{sec:approach}


Our proposed model uses segmentation labels as additional information to perform the image inpainting. Suppose we are given an incomplete input image $I_0$, our goal is to predict the complete image $I$, which is composed of two parts, $I_0$ and $I_R$, where $I_R$ is the reconstructed area of the missing hole. Here we also model the segmentation label map $S$ as the latent variable, which is similarly composed of $S_0$ and $S_R$, where $R$ represents the missing hole. Our whole framework contains three steps as depicted in Fig. \ref{fig:framework}. First, we estimate $S_0$ from $I_0$ using the state-of-the-art algorithm. Then the Segmentation Prediction Network (SP-Net) is used to predict $S_R$ from $I_0$ and $S_0$. Lastly $S_R$ is passed to the Segmentation Guidance Network (SG-Net) as the input to predict the final result $I$.

\subsection{Segmentation Prediction Network (SP-Net)}
\noindent\textbf{Network architecture} 
The goal of SP-Net is to predict the segmentation label map in the missing hole. The input to SP-Net is the 256x256xC incomplete label map $S_0$ as well as the 256x256x3 incomplete image $I_0$, where C is the number of label categories, and the output is the prediction of segmentation label map $S$ of size 256x256xC. Existing works have proposed different architectures of the generator, such as the encoder-decoder structure~\cite{pathak2016context} and FCN structure~\cite{iizuzuka2017globally}. Similar to \cite{iizuzuka2017globally}, the generator of SP-Net is based on FCN but replaces the dilation convolution layer with residual blocks, which could provide better learning capacity. Progressive dilated factors are applied to increase the receptive field and provide a wider view of input to capture the global structure of the image. To be more specific, our generator consists of four down-sampling convolution layers, nine residual blocks, and four up-sampling convolution layers. The kernel sizes are 7 in the first layer and last layer, and are 3 in other layers. The dilation factors of 9 residual blocks are 2 for the first three blocks, then 4 for another three blocks, and 8 for the last ones. The output channel for down-sampling layers and up-sampling layers are respectively 64, 128, 256, 512 and 512, 256, 128, 64, while they're all 512 for residual blocks. ReLU and Batch normalization layer is used between each convolution layer except the last layer which produces the final result. The last layer uses a softmax function to produce a probability map, which predicts the probability of the segmentation label for each pixel.

\noindent\textbf{Loss Functions}
Adversarial losses are given by discriminator networks to judge whether an image is real or fake and have been widely used since the emerging of GANs~\cite{goodfellow2014generative}. However, a single GAN discriminator design is not good enough to produce a clear and realistic result as it needs to take both global view and local view into consideration. To address this problem, we use the multi-scale discriminators similar to ~\cite{wang2017high} which have same network structure but operate at three different scales of image resolutions. Each discriminator is a fully convolutional PatchGAN~\cite{isola2016image} with 4 down-sampling layers followed by a sigmoid function to produce a vector of reak/fake predictions, where each value corresponds to a local patch in the original image. By the multi-scale application, the discriminators, i.e. $\{D_1, D_2, D_3\}$, take corresponding inputs that are down-sampled from the original image by a factor of 1, 2, 4 respectively, and are able to classify the global and local patches at different scales, which enable the generator to capture both global structure and local texture. More formally, the adversarial loss is defined as:
\begin{eqnarray}
\begin{split}
& \min_G \max_{D_1,D_2,D_3} \sum_{k=1,2,3} L_{GAN}(G,D_k) \\
& = \sum_{k=1,2,3} E[\log (D_k((S_0)_k,(S_{gt})_k) + \log (1-D_k((S_0)_k,(G(S_0)_k)].
\end{split}
\end{eqnarray}
Here $(S_0)_k$ and $(S_{gt})_k$ refer to the $k^{th}$ image scale of the input label map and ground truth respectively. 

Instead of the common reconstruction loss for image inpainting, we improve the loss by defining a perceptual loss, which is introduced by Gatys \etal~\cite{gatys2015neural}, and then widely used in many tasks aiming to improve the perceptual evaluation performance~\cite{johnson2016perceptual, dosovitskiy2016generating}. As the input image is a label map with C channels, we cannot apply the perceptual loss function on a pre-trained model, which usually takes an image as input. Therefore, a more reasonable way is to extract feature maps from multiple layers of both the generator and the discriminator to match the intermediate representations. Specifically, the perceptual loss is written as:
\begin{eqnarray}
L_{perceptual}(G) = \sum_{l=0}^n \frac{1}{H_l W_l} \sum_{h,w} || M_l \odot (D_k(S_0,S_{gt})^l_{hw} - D_k(S_0, G(S_0))^l_{hw}) ||_1. 
\end{eqnarray}
Here $l$ refers to the feature layers, and $\odot$ refers to the pixelwise multiplication. $M_l$ is the mask of the missing hole at layer $l$. Using the feature matching loss has been proposed in the image translation task~\cite{wang2017high}. Here we extend the design to incorporate the mask weight, which helps to emphasize more on the generation in the missing area. Another benefit comes from $l$ starting from 0 where the layer 0 is the input of the discriminator, which contains a reconstruction loss function in nature. 

Our full objective is then defined to combine both adversarial loss and perceptual loss:
\begin{eqnarray}
\min_G( \lambda_{adv}(\max_{D_1,D_2,D_3} \sum_{k=1,2,3} L_{GAN}(G,D_k))+ \lambda_{perceptual} \sum_{k=1,2,3}L_{perceptual}(G)),
\end{eqnarray}
where $\lambda_{adv}$ and $\lambda_{perceptual}$ control the rate of two terms. In our experiment, we set $\lambda_{adv}=1$ and $\lambda_{perceptual}=10$ as used in ~\cite{pathak2016context,wang2017high}.

\subsection{Segmentation Guidance Network (SG-Net)}
\noindent\textbf{Network architecture} 
The goal of SG-Net is to predict the image inpainting result $I$ of size 256x256x3 in the missing hole. It takes a 256x256x3 incomplete image $I_0$ jointly with the segmentation label map $S$ predicted by SP-Net as input. The SG-Net shares a similar architecture with SP-Net, with four down-sampling convolution layers, nine residual blocks and four up-sampling layers. Different from SP-Net, the last convolution layer uses a tanh function to produce an image with pixel value ranged at $[-1,1]$, which is then rescaled to the normal image value.

\noindent\textbf{Loss Functions}
Besides the loss functions in SP-Net, SG-Net introduces an additional perceptual loss to stabilize the training process. Traditional perceptual losses typically use VGG-Net and compute the $\ell_2$ distance on different feature layers. Recently \cite{zhang2018unreasonable} proposes to train a perceptual network based on AlexNet to measure the perceptual differences between two image patches and shows that AlexNet performs better to reflect human perceptual judgements. Here we extend the loss function by considering the local hole patch. The perceptual network computes the activations of the hole patches and sums up the $\ell_2$ distances across all feature layers, each scaled by a learned weight, which finally provides a perceptual real/fake prediction. Formally, the new perceptual loss based on AlexNet is defined as:
\begin{eqnarray}
L_{Alex}(G) = \sum_{l} \frac{1}{H_l W_l} \sum_{h,w} || w_l \circ (\Psi(G(I_0)_p)^l_{hw} - \Psi((I_{gt})_p)^l_{hw}) ||_2^2. 
\end{eqnarray}
Here p refers to the local hole patch, and $I_{0}$, $I_{gt}$ are the incompulete image and ground truth respectively. $\Psi$ is the AlexNet and $l$ is the feature layer. $w_l$ is the layer-wise learnt weight. With the benifit of this extra perceptual loss, the full loss function of SG-Net is defined as:
\begin{eqnarray}
\begin{split}
\min_G( & \lambda_{adv}\max_{D_1,D_2,D_3} \sum_{k=1,2,3} L_{GAN}(G,D_k)+ \\
& \lambda_{perceptual} \sum_{k=1,2,3}L_{perceptual}(G) + \lambda_{Alex} L_{Alex}(G)),
\end{split}
\end{eqnarray}
where we set $\lambda_{adv}=1$, $\lambda_{perceptual}=10$ and $\lambda_{Alex}=10$ in our experiment.


\section{Experiments}\label{sec:experiments}

\subsection{Experiment Setup}
We conduct extensive comparisons on two public datasets: Cityscapes dataset~\cite{cordts2016cityscapes} and Helen Face dataset~\cite{le2012interactive,smith2013exemplar}. Cityscapes dataset has 2,975 street view images for training and we use the validation set for testing, which consists of 500 images. Helen Face dataset has 2,000 face images for training and 100 images for testing. The fine annotations of the segmentation labels for both datasets are also provided for training. Cityscapes and Helen Face dataset are annotated with 35 and 11 categories respectively. To better capture the global structure of the street view, we map the 35 categories to 8 categories, which are road, building, sign, vegetation, sky, person, vehicle, and unlabeled otherwise.

To make fair comparisons with existing methods, we train images of size 256x256. For each image, we apply a mask with a single hole at random locations. The sizes of holes are between 1/8 and 1/2 of the image's size. To train the whole networks, we firstly use the state-of-the-art semantic segmentation method Deeplabv3+~\cite{deeplabv3plus2018} and fix its model parameters. Then we train the SP-Net and SG-Net separately for 200 epochs with linear learning rate decay in the last 100 epochs. Finally, we train the whole architecture in an end-to-end manner for additional 100 epochs. We train all our models on an NVIDIA Titan X GPU. The total training time for the two steps is around 2 days, and the inference is real-time.

\subsection{Comparisons}
For Cityscapes, We compare our method with 2 methods: PatchMatch~\cite{barnes2009patchmatch} and Globally-Locally consistent inpainting (GL)~\cite{IizukaSIGGRAPH2017}. PatchMatch is the state-of-the-art non-learning based approach, and GL is the recent work proposed by Iizuka \etal. We make the comparison in a random hole setting, and only GL applies Poisson Image Editing as post-processing as stated in ~\cite{IizukaSIGGRAPH2017}. For Helen Face dataset, we compare our method with Generative Face Completion (GFC)~\cite{li2017generative}.

\begin{figure*}[!h]
\centering
\small{}
\setlength{\tabcolsep}{1pt}
\begin{tabular}{cccccc}
 \includegraphics[width=.20\textwidth]{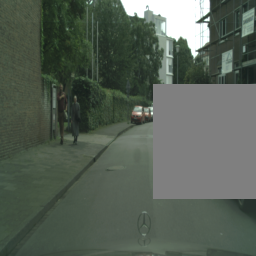}&
 \includegraphics[width=.20\textwidth]{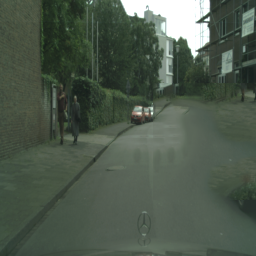}&
 \includegraphics[width=.20\textwidth]{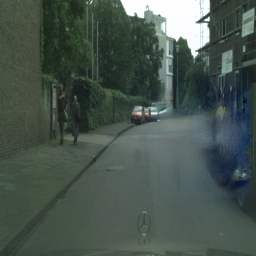}&
 \includegraphics[width=.20\textwidth]{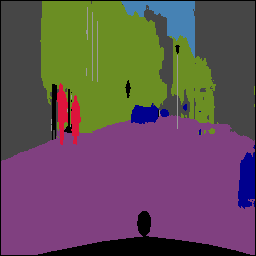}&
 \includegraphics[width=.20\textwidth]{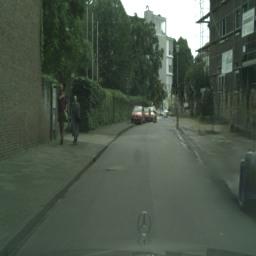}\\
 \includegraphics[width=.20\textwidth]{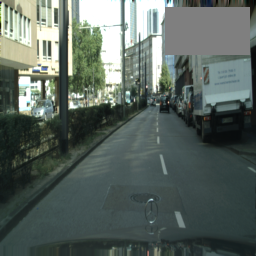}&
 \includegraphics[width=.20\textwidth]{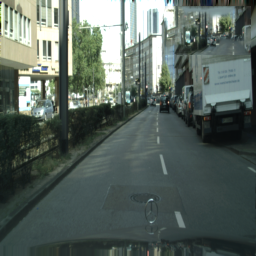}&
 \includegraphics[width=.20\textwidth]{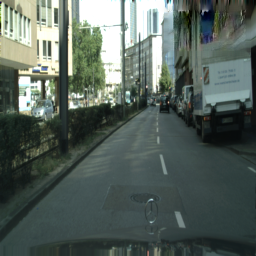}&
 \includegraphics[width=.20\textwidth]{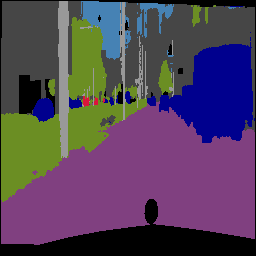}&
 \includegraphics[width=.20\textwidth]{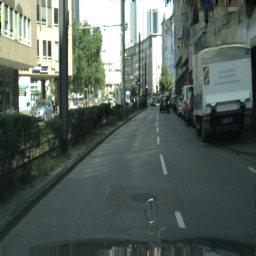}\\
 \includegraphics[width=.20\textwidth]{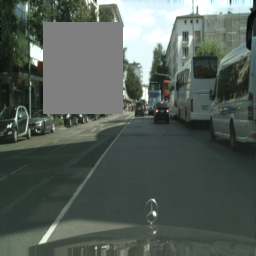}&
 \includegraphics[width=.20\textwidth]{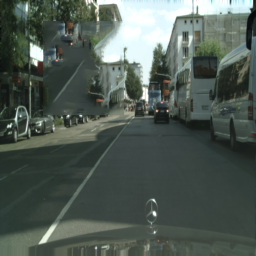}&
 \includegraphics[width=.20\textwidth]{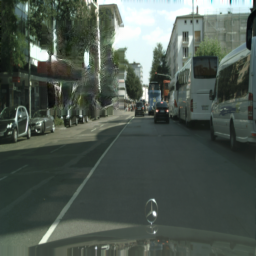}&
 \includegraphics[width=.20\textwidth]{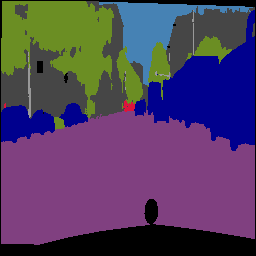}&
 \includegraphics[width=.20\textwidth]{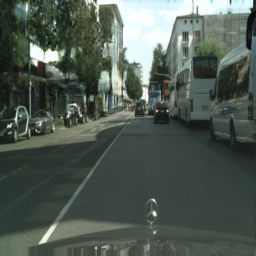}\\
 \includegraphics[width=.20\textwidth]{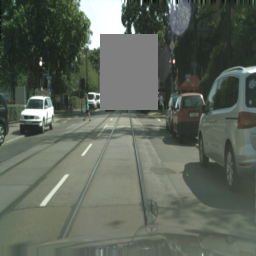}&
 \includegraphics[width=.20\textwidth]{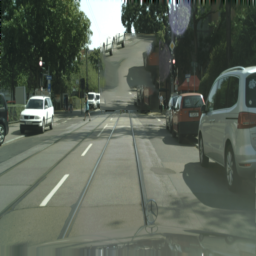}&
 \includegraphics[width=.20\textwidth]{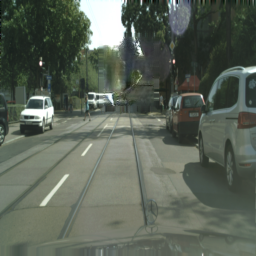}&
 \includegraphics[width=.20\textwidth]{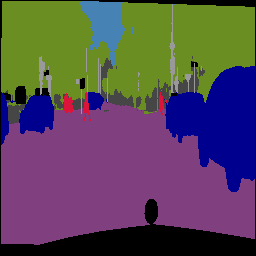}&
 \includegraphics[width=.20\textwidth]{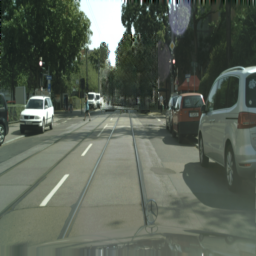}\\

\end{tabular}
\caption{Visual comparisons of Cityscapes results with random hole. Each example from left to right: input image, PatchMatch~\cite{barnes2009patchmatch}, GL~\cite{IizukaSIGGRAPH2017}, SP-Net output, and SG-Net output (our final result). All images have size $256\times 256$. Zoom in for better visual quality.}
\vspace{-10pt}
\label{fig:cityscapes}
\end{figure*}

\begin{figure*}[!h]
\centering
\small
\setlength{\tabcolsep}{1pt}
\begin{tabular}{cccccc}
 \includegraphics[width=.16\textwidth]{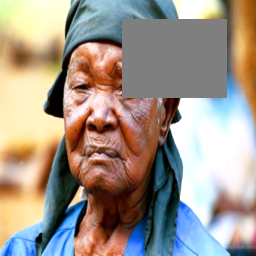}&
 \includegraphics[width=.16\textwidth]{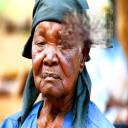}&
 \includegraphics[width=.16\textwidth]{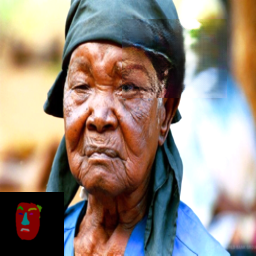}&
 \includegraphics[width=.16\textwidth]{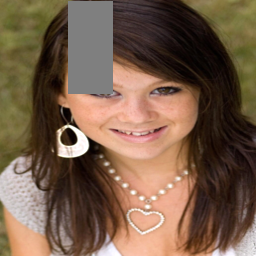}&
 \includegraphics[width=.16\textwidth]{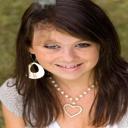}&
 \includegraphics[width=.16\textwidth]{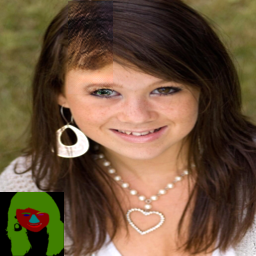}\\
 \includegraphics[width=.16\textwidth]{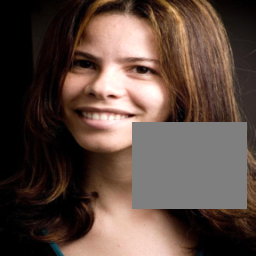}&
 \includegraphics[width=.16\textwidth]{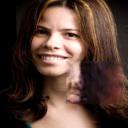}&
 \includegraphics[width=.16\textwidth]{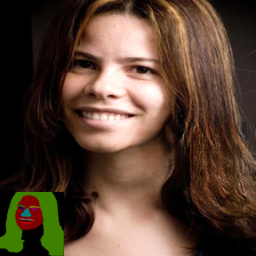}&
 \includegraphics[width=.16\textwidth]{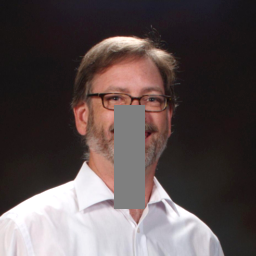}&
 \includegraphics[width=.16\textwidth]{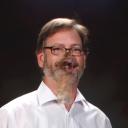}&
 \includegraphics[width=.16\textwidth]{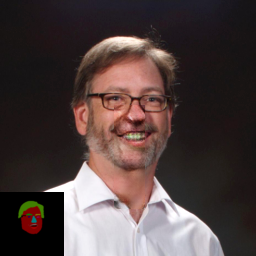}\\
 (a) & (b) & (c) & (d) & (e) & (f)\\
\end{tabular}
\caption{Visual comparisons of Helen Face Dataset results with random hole. Each example from left to right: input image, GFC~\cite{li2017generative}, and our result. All images have size $256\times 256$.}
\vspace{-10pt}
\label{fig:helenface}
\end{figure*}

\noindent\textbf{Qualitative Comparisons} 
Fig.~\ref{fig:cityscapes} shows the comparisons on Cityscapes dataset where images are randomly drawn from the test set. Cityscapes is a dataset of traffic view images with highly complicated global structures. We can see that PatchMatch could generate realistic details by patch propagation, but they often fail to capture the global structure or synthesize new contents in most scenarios. While GL could provide plausible textures which is coherent with the surrounding area, it could not handle the object shapes well and often predict unreasonable structures. As compared with GL, our SP-Net could focus on the task of shape prediction, and then pass the high-level semantic information as guidance for the generation step of SG-Net. This factorizes the inpainting task in a reasonable way, which enables the completion of different object shapes. For example, the masks in the second and third rows of Fig.~\ref{fig:cityscapes} contain an interaction of multiple object boundaries, such as the car, building, and tree. While GL only propagates the texture from the neighborhood of the holes and gives very blurry results, our SP-Net makes a more reasonable layout prediction and SG-Net recovers boundaries of the car and building very clearly. Furthermore, our method could also infer a part of a car even from a very small segmentation shape in the input and complete the wheel of the car in the first example of Fig.~\ref{fig:cityscapes}. Fig.~\ref{fig:helenface} shows the comparisons on Helen Face dataset. Since \cite{li2017generative}'s model deals with images of 128x128, we directly up-sample the results to 256x256 for comparison. For our results, the prediction of SP-Net is also shown at lower left corners. It can be seen that our method could also generate more realistic face inpainting results than GFC~\cite{li2017generative} which is specifically designed and trained for face completion, and this indicates the strong generalization ability of our segmentation based inpainting framework.

\noindent\textbf{Quantitative Comparisons} 
We make the quantitative comparisons between PatchMatch, GL, and our method. Here we report four image quality assessment metrics: $\ell_1$, $\ell_2$, SSIM~\cite{wang2004image}, and PSNR following works of \cite{pathak2016context,yang2017high}. Table \ref{table:numerical} shows the comparison results. It can be seen that our method outperforms the other methods on three out of the four metrics. For $\ell_2$, GL has slightly smaller errors than ours, but $\ell_2$ error is less capable to assess the perceptual quality than SSIM and PSNR, as it tends to average pixel values and award blurry outputs.

\begin{table}[h!]
\begin{center}

\resizebox{.80\textwidth}{!}{%
{\tiny
  \begin{tabular}{ l  c  c  c  c}
    \hline
    \textbf{Method} & \textbf{$\ell_1$ Error} & \textbf{$\ell_2$ Error} &  \textbf{SSIM} &  \textbf{PSNR} \\ \hline
    \emph{PatchMatch~\cite{barnes2009patchmatch}} & 641.3 & 169.3 & 0.9419 & 30.34\\ \hline
    \emph{GL~\cite{IizukaSIGGRAPH2017}} & 598.0 & \textbf{94.78} & 0.9576 & 33.57\\ \hline
    \emph{Ours} & \textbf{392.4} & 98.95 & \textbf{0.9591} & \textbf{34.26} \\ \hline
    \hline
  \end{tabular}}
  }
  \end{center}
  \caption{Numerical comparison on 200 test images of Cityscapes.}
  \vspace{-10pt}
  \label{table:numerical}
\end{table}

\noindent\textbf{User Study} 
To better evaluate our methods from the perceptual view of people, we conduct a user study on the Cityscapes dataset to make comparisons. We ask 30 users for perceptual evaluation, each with 20 subjective tests. At every test, users are shown the input incomplete image and are asked to compare the results of PatchMatch, GL, and ours. Among 600 total comparisons, the user study shows that our results receive the highest score 70.8\% of the time. As compared with PatchMatch, our results are overwhelmingly better 96.2\% of the time. Comparing with GL, our results are perceptually better 71.3\% of the time, and are ranked the same 16.3\% of the time.

\subsection{Analysis}

\begin{figure}[!h]
\centering
\small
\setlength{\tabcolsep}{0.5pt}
\begin{tabular}{ccccc}
  \includegraphics[width=.20\textwidth]{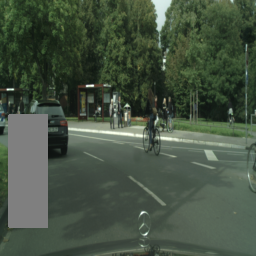}&
  \includegraphics[width=.20\textwidth]{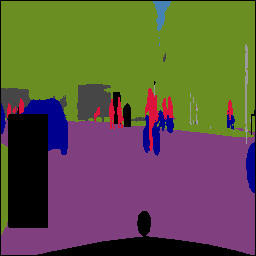}&
  \includegraphics[width=.20\textwidth]{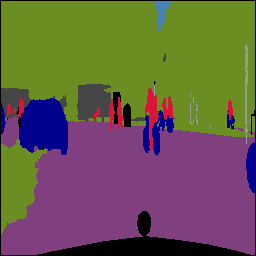}&
  \includegraphics[width=.20\textwidth]{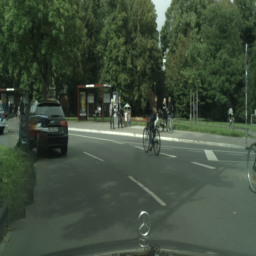}&
  \includegraphics[width=.20\textwidth]{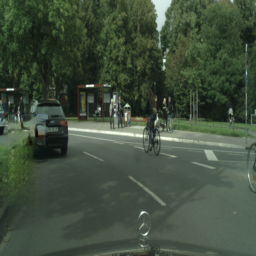}\\
  (a) & (b) & (c) & (d) & (e)\\
\end{tabular}
\caption{Ablation study. (a) Input image with missing hole. (b) Deeplabv3+~\cite{deeplabv3plus2018} output. (c) SP-Net result. (d) SG-Net result. (e) Baseline result. The size of images are 256x256.}
\label{fig:ablation}
\vspace{-10pt}
\end{figure}

\noindent\textbf{Ablation Study} 
Our main motivation is to introduce segmentation label map as intermediate guidance to provide high-quality inpainting results. To justify this framework, we show the intermediate results of our method at each step and compare our result to the baseline result. Here the baseline result refers to the single SG-Net, which only takes the incomplete image as input and doesn't have any other conditions. We train both methods in the same setting with 200 epochs, and show the comparison in Fig.~\ref{fig:ablation}. We can see that Deeplabv3+ provides an accurate segmentation label map, and SP-Net could make a reasonable prediction. SG-Net then inpaints the missing area based on the output of SP-Net, which generates sharp and realistic details. Comparing with our method, the baseline result is very blurred, especially along the boundaries of the vegetation and the car.

\begin{figure}[!h]
\centering
\small
\setlength{\tabcolsep}{0.5pt}
\begin{tabular}{cccccc}
  \includegraphics[width=.16\textwidth]{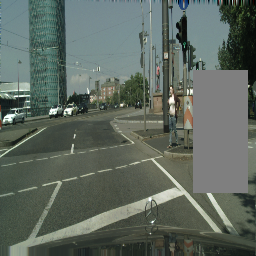}&
  \includegraphics[width=.16\textwidth]{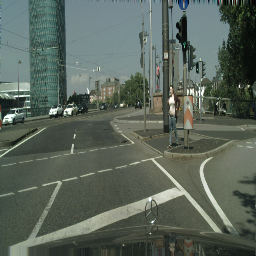}&
  \includegraphics[width=.16\textwidth]{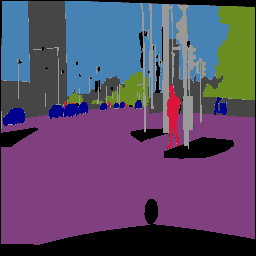}&
  \includegraphics[width=.16\textwidth]{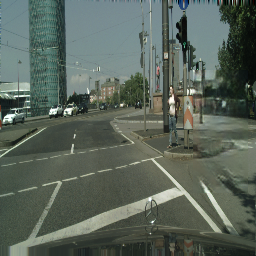}&
  \includegraphics[width=.16\textwidth]{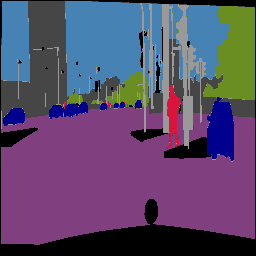}&
  \includegraphics[width=.16\textwidth]{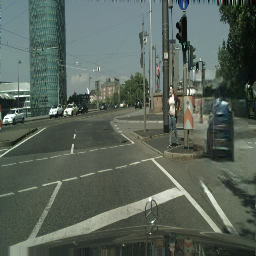}\\
  (a) & (b) & (c) & (d) & (e) & (f)\\
\end{tabular}
\caption{Interactive Editing. (a) Input image with missing hole. (b) Ground truth. (c) First label map. (d) Inpainting result based on (c). (e) Second label map. (f) Inpainting result based on (e). The size of images are 256x256.}
\label{fig:interactive}
\vspace{-10pt}
\end{figure}

\noindent\textbf{Interactive Editing} 
Our segmentation based framework allows us to perform interactive editing on the inpainting task to give multi-modal predictions for each single input image. Specifically, when we are given an incomplete image as input, we don't know the ground truth layout in the missing hole. However, we could make interactive editing on the segmentation map in the mask, such as following the ground truth label to guide the inpainting (Fig.~\ref{fig:interactive}c), or add more components to the hole, \eg a car (Fig.~\ref{fig:interactive}e). While both label maps are reasonable, SG-Net could provide multi-modal outputs based on different conditions (Fig.~\ref{fig:interactive}df).

\section{Conclusion}
\label{sec:conclusion}

In this work, we propose a novel end-to-end learning framework for image inpainting. It is composed of two distinct networks to provide segmentation information and generate realistic and sharp details. We have observed that segmentation label maps could be predicted directly from the incomplete input image, and provide important guidance for the texture generation in the missing hole. Our method also allows an interactive editing to manipulate the segmentation maps and predict multi-modal outputs. We expect that these contributions broaden the possibilities for image inpainting task and could be applied to more image editing and manipulation applications.



\bibliography{egbib}
\end{document}